\documentclass[10pt]{article}
\usepackage{verbatim} 
\usepackage{amsmath,graphicx}
\usepackage[normalem]{ulem}
\usepackage[preprint]{spconf}

\usepackage{ifthen}
\usepackage{fancyhdr}
\title{DEEP LEARNING FOR PROMINENCE DETECTION IN CHILDREN'S READ SPEECH}

\name{Mithilesh Vaidya \qquad Kamini Sabu \qquad Preeti Rao}
\address{Department of Electrical Engineering,\\
  Indian Institute of Technology Bombay, Mumbai, India}

\begin{document}

\maketitle

\begin{abstract}
The detection of perceived prominence in speech has attracted approaches ranging from the design of linguistic knowledge-based acoustic features to the automatic feature learning from suprasegmental attributes such as pitch and intensity contours. We present here, in contrast, a system that operates directly on segmented speech waveforms to learn features relevant to prominent word detection for children's oral fluency assessment. The chosen CRNN (convolutional recurrent neural network) framework, incorporating both word-level features and sequence information, is found to benefit from the perceptually motivated SincNet filters as the first convolutional layer. We further explore the benefits of the linguistic association between the prosodic events of phrase boundary and prominence with different multi-task architectures. Matching the previously reported performance on the same dataset of a random forest ensemble predictor trained on carefully chosen hand-crafted acoustic features, we evaluate further the possibly complementary information from hand-crafted acoustic and pre-trained lexical features.

\end{abstract}

\keywords{word prominence, phrase bounday, multi-task learning, children's speech prosody}

\section{INTRODUCTION}\label{sec:intro}
The prosodic structure of speech carries important information in terms of the syntax and the meaning, both of which are critical to a listener's ease of comprehension of the spoken message~\cite{1983bock_MC_comprehension,2017maastricht_is_intonation,2017levis_pslt_infoStructure}. Phrase boundaries embed sentence syntax through word grouping while prominence or emphasis on specific words signals new information or highlights a contrast.  
Given their importance in applications requiring speech understanding such as scoring of spoken language fluency and text-to-speech synthesis, the automatic detection of perceived prominence and phrase boundaries has attracted continuous research efforts. Prominence or word stress, the focus of the current work, has proved more challenging of the two.

Prominence is perceived by a listener when a word stands out of its local context in one or more of the suprasegmental attributes such as duration, F0, intensity and spectral shape \cite{2010breen_LCP_infostructwithacoustics}. The local context itself refers to the phones and syllables within the word as well as a neighborhood of up to several words. Prosody perception, however, is influenced not only by the low-level acoustic cues but also top-down expectations from lexico-syntactic information~\cite{2010cole_LP_prominence,2018baumann_JP_prominPoSgerman}. The precise combination and relative importance of the cues depends on the speaker, language and speaking style as also on the listener. Traditionally, various aggregates of the sampled acoustic parameters across the word segment including mean and variance, contour shape descriptors, and differences in these quantities across neighboring words comprise word-level prosodic features~\cite{2009rosenberg_thesis_prosodiceventclassification, 2012mishra_is_wordprominence}. These features are then used to train a conventional supervised classifier for the automatic detection, possibly in combination with lexico-syntactic information~\cite{2014christodoulides_avanzi_is_prominence,2015black_narayanan_is_nonnativeaccent}. In our own recent work, we used feature selection on a large set of features, computed across the distinct suprasegmental attributes of speech, in a random forest ensemble predictor to derive a compact set of interpretable features for speaker-independent boundary and prominence detection on a children's oral reading dataset~\cite{2021sabu_CSL_prosodicevent}. It was observed that apart from the expected pitch, duration and intensity based aggregates, the acoustic cues to prominence included a number of spectral shape functionals while the phrase boundary prediction was dominated by pause based features. With the search space for such `hand-crafted' features being very large, however, the process can miss potentially important features. Further, the pre-selected context windows used in such analyses make it difficult to exploit the long and variable time scale of prosodic relationships across an utterance in any comprehensive manner. The potential for deep learning solutions has therefore been recognized for some time but incorporated successfully in the prominence detection task only more recently, as briefly reviewed next.

Rosenberg et al.~\cite{2015rosenberg_is_rnn} used a large number of acoustic-prosodic features and aggregates at word level derived from their previous AuToBI work \cite{2009rosenberg_thesis_prosodiceventclassification,2010rosenberg_is_AuToBI} in a BiRNN classifier where the word sequence context was learned over that explicitly provided in the feature vector. They observed a small improvement ($<$ 1\% absolute) in boundary and pitch accent detection over a baseline conditional random forest classifier.
Wu et al.~\cite{2019wu_ssps_pitchaccent} also used similar aggregated acoustic features with an LSTM to find an improvement over the use of an SVM classifier. Lin et al.~\cite{2020lin_is_prosodicevents} used a hierarchical BLSTM network to aggregate features across phone, syllable and word to model contextual information at multiple granularities in the joint detection of boundaries and prominence. 

In a departure from the above pre-computed word-level features, Stehwien et al.~\cite{2017stehwein_is_CNNforprosodicevent,2020stehwien_SC_CNNforprosodicevents} used CNN on sampled acoustic parameters (energy, F0, loudness, voicing probability, zero crossing rate and harmonic-to-noise ratio) together with a context window of two neighbouring words to optimally learn the word-level aggregated features. The max-pooled CNN feature maps are directly classified with a softmax layer. With word position indicators provided in the input segment, they report an improvement of 1-3\% absolute over Rosenberg~\cite{2009rosenberg_thesis_prosodiceventclassification} on lexical stress and phrase boundary detection on the BURNC corpus, with speaker-independent scenarios being more challenging. 
Both local acoustic features and longer, more global contexts spanning several words and possibly different sentences across the utterance are important in the perception of prominence. Hence, architectures combining low-level feature aggregation with sequence models were realized with the same contour-learned features input to an LSTM classification layer~\cite{2020nielsen_emnlp_prominenceCNN, 2018zhang_iscslp_emphasisinDialogue}.


Feature learning via end-to-end neural network systems trained on speech waveforms is being increasingly viewed as the optimal approach to complex classification tasks~\cite{2019rajan_SPL_conflictnet, ravanelli2018speaker, 2020fritsch_icassp_sleepiness}. Such systems have achieved performances close to, but not always exceeding, those of classifiers with task-specific hand-crafted features. This hints at a need for the introduction of reasonable constraints, or additional information, in the network architectures especially in the widely encountered data-constrained scenarios. 
In this work, we explore precisely such variations for the prominence detection task from segmented speech waveforms starting from a straightforward CRNN model. This is the first case of prosodic event detection from speech waveforms that we are aware of. 

The first variation involves replacing the CNN layer at the input with bandpass constrained, but tunable, filters motivated by the traditionally used mel filterbanks emulating the low-level auditory processing~\cite{ravanelli2018speaker}. SincNet has been applied in frame-level speaker identification where its hyperparameters  have been found to be critical, although sometimes counter-intuitive, to achieved task accuracy~\cite{oneatua2021revisiting}.
Next, we try to exploit the linguistic association between phrase boundaries and prominent words with multi-task learning. 
The presence of phrase-finality increases the perceived prominence of the word ~\cite{2017roy_cole_mahrt_LP_prosodicevent,2020bishop_JP_prominperception} and can potentially contribute to the feature representation for prominence. 
Recent work on the joint prediction of boundary and prominence is promising, with the boundary predictions computed from the final layer output of a 3-layer BLSTM network while prominence predictions are made at the penultimate layer~\cite{2020lin_is_prosodicevents}. In another attempt, prosodic event classification is viewed as a 4-class problem~\cite{2020suni_sp_jointevent}. 
Developing the above theme further, we explore alternate multi-task architectures for prominence detection that incorporate information about the (typically more reliably predicted) phrase boundary status of the word in distinct ways.

Our performance baseline is previous work on the same dataset and task in which a random forest ensemble predictor uses highly tuned hand-crafted features~\cite{2021sabu_CSL_prosodicevent}. We also examine the possibly complementary information of the hand-crafted features. Finally, given the importance of lexical information in prominence detection~\cite{2019talman_CORR_BERT, 2018stehwien_sp_wordEmbedding}, we report the combination with pretrained word embeddings.

\section{DATASET AND TASK}
\label{sec:dataset}
The children's oral reading dataset used in this work comprises recordings of grade-appropriate text read aloud by selected middle school students with reasonable word decoding ability in English (as second language) but widely varying levels of prosodic skill~\cite{2021sabu_CSL_prosodicevent}. The individual utterances are story paragraphs comprising between 50-70 words, each word labeled separately for the presence/absence of prominence and phrase boundary by 7 naive listeners using the RPT methodology~\cite{2017cole_mahrt_roy_CSL_rpt}. This is reduced to a net rating per word based on number of votes (out of 7), further scaled down to range 0-1, to obtain the `degree' of prominence (boundary) per word. The dataset contains 41,286 words across 790 utterances by 35 speakers, recorded at 16 kHz sampling rate. The utterances are available segmented at word level by forced alignment with the manual transcript.

The waveform for each segmented word is padded to the maximum size in the dataset, which is approximately 1.79 seconds (or 28,660 samples). A silent pause before the word (with duration limited to 500 ms if higher) is included in the waveform segment. 
The dataset is split into three equal folds with no speaker overlap for 3-fold cross-validation based testing. The hyperparameters are tuned with 4-fold CV on the train split. The 4 trained models are used for inference and their predictions are averaged to generate results on the corresponding unseen test set. The results for the prominence degree prediction are reported in terms of Pearson correlation between the predicted output and the degree of prominence from the RPT rater votes which serves as ground truth. We report the mean and standard deviation across the three test folds.

\section{MODEL}

The input to our prominent word detection network is the sequence of word-level waveform segments. 
We build up our model from a basic CRNN operating on waveform segments to more complex architectures that incorporate additional information. 

\begin{figure}[t]
  \centering
  \includegraphics[width=8cm]{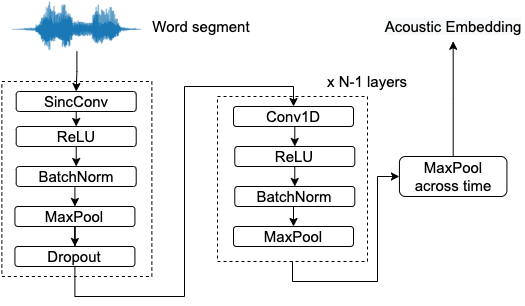}
  \caption{CNN component of the model, where the first layer is replaced with a Sinc layer.}
  \vspace{-0.2in}
  \label{fig:cnn}
\end{figure}

\subsection{Waveform CRNN}
\label{ssec:crnn}

As the name suggests, CRNN consists of a convolutional neural network, followed by a recurrent neural network. The CNN layers output a fixed-dimensional feature embedding, while the RNN processes the sequence of embeddings.
As in some recent high-level speech classification systems, our CNN is composed of multiple layers, where each layer consists of 1D convolution with batch normalization, 
ReLU activation and max pooling~\cite{2019rajan_SPL_conflictnet}. After N CNN layers (where N is a hyperparameter), the CNN output is max-pooled across time to get a fixed dimensional embedding for each word.

With a motivation to improve the input representation with meaningful parameter constraints in the first convolutional layer, we experiment with Sinc filters as shown in Figure~\ref{fig:cnn}. The constrained band-pass Sinc filters are expected to not only improve performance but also speed up training. 
The hyperparameters of the Sinc layer (number of filters, window width and stride) are tuned on the data and task. 

Prominent words stand out through the acoustic changes in the local context~\cite{2016kakouros_rasanen_SC_3pro}.  Prominence detection systems typically include the context explicitly during the feature extraction. In deep learning, 
the word context that is critical in speech prosody perception is modelled by the sequence classifier GRU. The relevant local context is therefore learned from the sequence of words spanning the entire utterance. 
At each time step, the GRU takes as input a fixed-dimensional vector corresponding to each word. This vector could be the acoustic embedding extracted from the CNN, word-level features such as lexical features or a concatenation of the two. The output of the, possibly bidirectional, GRU at each time step is fed to a dense network consisting of two fully-connected (FC) layers.
The first FC layer, consisting of 128 neurons, is followed by ReLU activation and a dropout layer. The second FC layer has one neuron, which is followed by sigmoid activation to get the final score.
Mean Squared Error (MSE) between model prediction and ground truth is minimised.

\subsection{Multi-task learning extension}

\begin{figure}[t]
  \centering
  \includegraphics[width=8cm]{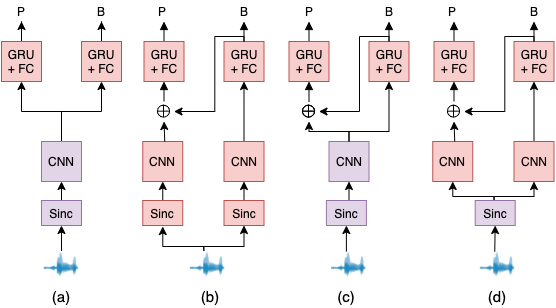}
  \caption{MTL architectures: (a) Sharing paradigm in which a common CNN feature extractor is followed by separate GRU heads for prediction of prominence (P) and boundary (B). (b) Conditioned MTL with separate CNNs for P and B predictions. (c) Shared CNN combined with conditioning by boundary prediction. (d) Conditioned MTL with only the Sinc layer shared. Note that $\oplus$ denotes concatenation. GRU blocks also include two fully-connected layers at the output for final prediction as discussed in Section \ref{ssec:crnn}}
  \label{fig:mtl}
  \vspace{-0.2in}
\end{figure}

As discussed in Section~\ref{sec:intro}, phrase boundary detection is a closely related task which can bring in complementary information to boost the prominence prediction performance. Given the different ways in which the relevant auxiliary task can be incorporated in the training, we experiment with two distinct multi-task learning (MTL) paradigms: parameter sharing and conditioning~\cite{kelz2019multitask}.

In parameter sharing, the two tasks share the initial feature extraction through shared CNN layers, while the subsequent layers (GRU and Dense) responsible for the final classification are different, as depitced in 
Figure \ref{fig:mtl}(a). This sharing of parameters at the feature extraction stage prevents overfitting and hence improves generalisation. On the other hand, in conditioned MTL, we have a separate network branch for each task with the final boundary prediction provided as an additional input to the GRU of the prominence prediction task. 
In this case, the CNN feature extractor can be shared (Figure~\ref{fig:mtl}(b)) or separate (Figure~\ref{fig:mtl}(c)). 
Based on an understanding of the role played by the Sinc layer in emulating low level auditory feature extraction, we explore a new architecture in which the Sinc layer is shared across the two tasks but the subsequent CNN, GRU and Dense layers are separate, as in Figure~\ref{fig:mtl}(d). 

For MTL, the final loss is a convex combination of the prominence MSE loss ($L_{prominence}$) and phrase boundary MSE loss ($L_{boundary}$) i.e.
\begin{equation}
    \label{eq}
    L_{total} = \alpha L_{prominence} + (1 - \alpha)L_{boundary}
\end{equation}
where $\alpha$ is a hyperparameter which controls the trade-off between performance on the main task and the auxiliary task. For the single-task experiments, $\alpha = 1$.

\subsection{Hyperparameters}
\label{ssec:hyperparam}
The hyperparameters for the basic CNN model are tuned for performance on the validation set. These were varied in the following ranges: number of layers, N (2-8), number of filters (16-128), CNN kernel width (7-151), pooling width (2-4) and stride (1 and 2). We found the optimal configuration to be: 4 layers, each consisting of 32 filters of kernel width 51, stride 1 and max pooling with kernel size of 3.

For the GRU, it was found that a 3-layer, 256-dimensional bidirectional GRU with dropout of 0.5 at each output layer (except the last layer) consistently gave the best performance. The hyperparameters of the GRU and the subsequent FC layers are fixed for all experiments.

For training, Adam~\cite{kingma2017adam} optimizer is used with a learning rate of 0.001. Batch size is set to 64 and the model is trained on a single NVIDIA GeForce GTX 1080. Early stopping is used on the validation set with patience set to 12 epochs. A single round of training and evaluation on our dataset took about 6 hours.

\section{EXPERIMENTS}
Prominence prediction is evaluated at the word level via its correlation with the ground-truth degree of prominence. We report here the test data performances for the different models in terms of the mean and standard deviation across the 3 test folds discussed in Section~\ref{sec:dataset}. 

\subsection{Single-task architectures}


\begin{table}[]
\caption{Performance of different architectures in single-task learning. A34 refers to the 34 hand-crafted features of~\cite{2021sabu_CSL_prosodicevent}. 
Number of filters in all CNN and Sinc layers is 32 while pool size is 3. For layers 2, 3 and 4, the hyperparameters are fixed as in Section~\ref{ssec:hyperparam} while Layer 1 variations are reported here in terms of Pearson correlation (s.d. $<$ 0.01).}
\centering
\label{tab:acoustic}
\begin{tabular}{|c|c|c|c|c|}
\hline
No. & Input & Acoustic & Layer 1 & Pearson\\
 & & model & (type, width, stride) & correl.\\ \hline \hline
1. & A34 & RFC & - & 0.696 \\
2. & A34 & GRU & - & 0.726 \\ \hline
3. & Wav & \begin{tabular}[c]{@{}l@{}}CRNN\end{tabular} & Standard, 51, 1 & 0.692 \\
4. & Wav & \begin{tabular}[c]{@{}l@{}}CRNN\end{tabular} & Sinc, 51, 1 & 0.712 \\
5. & Wav & \begin{tabular}[c]{@{}l@{}}CRNN\end{tabular} &Sinc, 31, 2 & 0.721 \\ \hline
6. & \begin{tabular}[c]{@{}l@{}}A34 +\\ Wav\end{tabular} & \begin{tabular}[c]{@{}l@{}}CRNN\end{tabular} & Sinc, 31, 2 & 0.735 \\ \hline
\end{tabular}
\vspace{-0.2in}
\end{table}


We start with reporting the performance of the baseline system that uses a set of 34 word-level acoustic-prosodic features computed on the acoustic contours of pitch, intensity and spectral shape versus time as well as various segmental durations including pauses. The final set of 34 features (termed `A34' here) is eventually obtained by two stages of feature selection in a random forest ensemble predictor~\cite{2021sabu_CSL_prosodicevent}. We also test the same A34 set of features with a bidirectional GRU classifier to study whether the implicit `learning' of word context can bring benefits. The results appear in Table~\ref{tab:acoustic}. We note that the performance shows a clear improvement with the sequence classifier, indicating the value of learned context over that explicitly represented within the A34 feature computations. 

Next, we evaluate the performance of our CRNN model operating directly on the speech waveform. We note, from Table~\ref{tab:acoustic}, a rise in the Pearson correlation with the Sinc layer replacing the (unconstrained) first convolutional layer but keeping the number of filters and filter widths and stride unchanged. A reduction of the Sinc filter widths to 31 samples (2 ms) gave a further improvement, especially when the stride was concurrently changed to 2 samples from 1 sample. This is consistent with the observations of previous work that smaller filter widths in the Sinc layer are superior in the context of speaker recognition~\cite{oneatua2021revisiting}. While this seems counter-intuitive given that auditory filter impulse responses at lower centre frequencies are of duration well over 10 ms, the reduced frequency resolution due to the apparent truncation does not seem to harm the performance. It is also possible that the slightly higher stride (2 samples) helps to counteract the shortened filter widths to some extent.  
It is encouraging to see that the gap between waveform-learned and the hand-crafted A34 features almost closing with this tuned Sinc version. 
To check for any complementarity in the two representations, we concatenate the A34 features with the 32-dimensional CNN embedding in the final column of the table to obtain a performance that exceeds that of either. However, ablation studies involving different suprasegmental attributes underlying the A34 features did not reveal any specific contribution as dominant. This indicates the future potential for better waveform-based feature learning, possibly with a larger training dataset.

\subsection{Multi-task architectures}

\begin{table}[]
\caption{Performance of various multi-task learning architectures and additional features. (Pearson correlation s.d. $<$ 0.01)}
\label{tab:mtl}
\centering
\begin{tabular}{|c|c|c|}
\hline
No. & MTL variant & Pearson\\
 & and additional features & correl. \\ \hline \hline
1. & Tuned Sinc (without MTL) & 0.721 \\ \hline
2. & Fig 2(a) & 0.726 \\ 
3. & Fig 2(b) & 0.727 \\ 
4. & Fig 2(c) & 0.724 \\ 
5. & Fig 2(d) & 0.740 \\ \hline
6. & Fig 2(d) + A34/A27 & 0.757 \\
7. & Fig 2(d) + A34/A27 + GloVe & 0.813 \\ \hline
\end{tabular}
\vspace{-0.2in}
\end{table}
The MTL experiments reported in Table~\ref{tab:mtl} were carried out after first tuning $\alpha$. After a preliminary grid search, we found that the best performance across configurations is obtained when $\alpha$ is set to 0.95 in equation~\ref{eq} (after scaling the MSE of each to bring them into the same range).  
From the correlations reported in Table~\ref{tab:mtl}, we note that neither of conditioning or shared CNN layers is better than the other. There is no clear improvement over single-task learning seen either. On the other hand, an increase in performance is seen with conditioned MTL when only the Sinc layer is shared and the other CNN layers remain task-dependent. This is consistent with our expectation that the lowest level features extracted from the input waveform correspond to the basic suprasegmental attributes fundamental to all prosodic event detection. Therefore the parameters of the constrained convolutional layer, that is the Sinc layer, get even better trained in the multi-task set-up. 

In row 6 of Table~\ref{tab:mtl}, we report performance on the concatenation of the hand-crafted acoustic features with the generated CNN embeddings at the GRU input. Similar to the A34 features for prominence, A27 refers to an optimal set of 27 word-level hand-crafted features for the task of boundary prediction obtained in~\cite{2021sabu_CSL_prosodicevent}. We expect the boundary task to benefit from the hand-crafted boundary detection features (A27) and therefore influence the prominence prediction performance. 
The observed improvement in performance confirms the presence of complementary acoustic information not captured by the purely waveform-based architecture.

\subsection{Incorporating lexical features}
Prominence is linked to the text syntax and semantics with content words such as proper nouns expected to receive prominence most of the time, followed by adjectives, nouns, adverbs, and verbs respectively~\cite{2018baumann_JP_prominPoSgerman}.
Lexical features in the form of GloVe embeddings can implicitly capture such parts-of-speech information and have been found to help in the context of prosodic event detection~\cite{2018stehwien_sp_wordEmbedding}. To explore the usefulness of lexical information for our task, we test two popular word embeddings: GloVe~\cite{pennington2014glove} and BERT~\cite{devlin-etal-2019-bert}. In the present work, we extract 100-dimensional GloVe embeddings pre-trained on Wikipedia using the gensim package. 
The embedding is passed through a dropout layer followed by a linear layer whose dimension is a hyperparameter~\cite{2018stehwien_sp_wordEmbedding}. After tuning, we found that a dropout layer of probability 0.3 and a fully-connected layer of dimension 300 gave the best performance. Although BERT has replaced GloVe embeddings in a wide variety of NLP tasks and demonstrated good performance for prominence detection~\cite{2019talman_CORR_BERT}, it did not give any clear improvement over GloVe in our task. This could be attributed to very simple story texts without semantic ambiguities that may benefit from contextualized embeddings. 
We note a big jump in performance in the final row of Table~\ref{tab:mtl} where the GloVe features are concatenated with the corresponding CNN embedding and hand-crafted features in each branch of the best MTL model, emphasising the importance of lexical features. 
Although this is at odds with the expectation that beginning readers do not necessarily realize prominence correctly, it supports the important role of the top-down expectations in raters' perceptions.


\section{CONCLUSIONS}

In this paper, we attempt to replace hand-engineered features for prominent word detection with deep learning models operating on the speech waveform. Our results indicate that it is challenging to surpass the performance of hand-crafted features computed from across the prior extracted suprasegmental contours of speech essential to prosody realization, at least with moderate sized training datasets. However, some human speech and audition motivated constraints such as Sinc based convolution at the lowest feature extraction stage can improve the performance of deep learning models. The optimal hyperparameters for the Sinc filters turned out to correspond to low strides and low widths (high time resolution and low frequency resolution), confirming previous findings in the speaker identification context. Hand-crafted features further enhance the performance, indicating  opportunities for further careful constraining and tuning of the deep learning models.

The use of boundary event labeled speech seems to aid the task of prominent word detection, as expected from the linguistic and acoustic association between the two. A multi-task architecture in which prominence prediction is conditioned on the concurrently predicted boundary and where the Sinc convolution filters are shared across the two tasks obtained the best performance. Finally, we  investigate the contribution of pre-trained lexical embeddings to prominence prediction to find a large increase in the performance. This indicates that the lexical identity of words guides the top-down expectations of raters, even in the case of not-so-proficient beginning readers.


In future work, we plan to explore more sophisticated CNN architectures, attention-based sequence modelling for better context representation and alternate methods to fuse acoustic and lexical information.



\bibliographystyle{IEEEtran}

\bibliography{template}

\end{document}